\documentclass[letterpaper, 10 pt, conference]{ieeeconf}  %

\IEEEoverridecommandlockouts                              %

\usepackage{amsmath} %
\usepackage{amssymb}  %
\usepackage{bm}

\usepackage[pdftex]{graphicx}

\usepackage{algorithm}
\usepackage{algorithmicx}
\usepackage{algpseudocode}
\usepackage{xcolor}
\usepackage{xspace}

\newcommand*{\algname}{\mbox{SINGER}\xspace}

\title{\LARGE \bf
SINGER: An Onboard Generalist Vision-Language Navigation Policy for Drones
}

\author{Maximilian Adang,$^{1}$ JunEn Low,$^{2}$ Ola Shorinwa,$^{1}$ and Mac Schwager$^{1}$%
\thanks{$^{1}$Department of Aeronautics and Astronautics, Stanford University, Stanford, CA 94404, USA {\tt\small \{madang, shorinwa, schwager\}@stanford.edu}}
\thanks{$^{2}$Department of Mechanical Engineering, Stanford University, Stanford, CA 94404, USA {\tt\small jelow@stanford.edu}}%
\thanks{*The first author was supported on an NDSEG fellowhsip. Toyota Research Institute provided funds to support this work.}%
}

\begin{document}

\maketitle
\thispagestyle{empty}
\pagestyle{empty}

\begin{abstract}
Large vision-language models have driven remarkable progress in open-vocabulary robot policies, e.g., generalist robot manipulation policies, that enable robots to complete complex tasks specified in natural language. Despite these successes, open-vocabulary autonomous drone navigation remains an unsolved challenge due to the scarcity of large-scale demonstrations, real-time control demands of drones for stabilization, and lack of reliable external pose estimation modules. 
In this work, 
we present \algname for language-guided autonomous drone navigation in the open world using only onboard sensing and compute. To train robust, open-vocabulary navigation policies, \algname leverages three central components: (i) a photorealistic language-embedded flight simulator with minimal sim-to-real gap using Gaussian Splatting for efficient data generation, (ii) an RRT-inspired multi-trajectory generation expert for collision-free navigation demonstrations, and these are used to train (iii) a lightweight end-to-end visuomotor policy for real-time closed-loop control.
Through extensive hardware flight experiments, we  demonstrate superior zero-shot sim-to-real transfer of our policy to unseen environments and unseen language-conditioned goal objects. When trained on $\sim$700k-1M observation action pairs of language conditioned visuomotor data and deployed on hardware, \algname outperforms a velocity-controlled semantic guidance baseline by reaching the query $\mathbf{23.33\%}$ more on average, and maintains the query in the field of view $\mathbf{16.67\%}$ more on average, with $\mathbf{10\%}$ fewer collisions.

\end{abstract}

\section{Introduction}\label{sec:intro}

Everyday, humans demonstrate notable semantic and physical understanding of their environments. For example, given a task to go to a specified location, a person relatively easily transforms the language instruction into a physical goal location using semantic cues and navigates to the desired location, safely avoiding collisions.
Although autonomous drones excel at agile flight, they are often limited to controlled environments with pre-specified goal locations. In this work, we ask the question: ``\textit{Can we train a vision-language drone navigation policy to reach previously unseen goal objects in a previously unseen environment using only on board sensing and compute?}"

\begin{figure}
    \centering
    \includegraphics[width=1.0\linewidth]{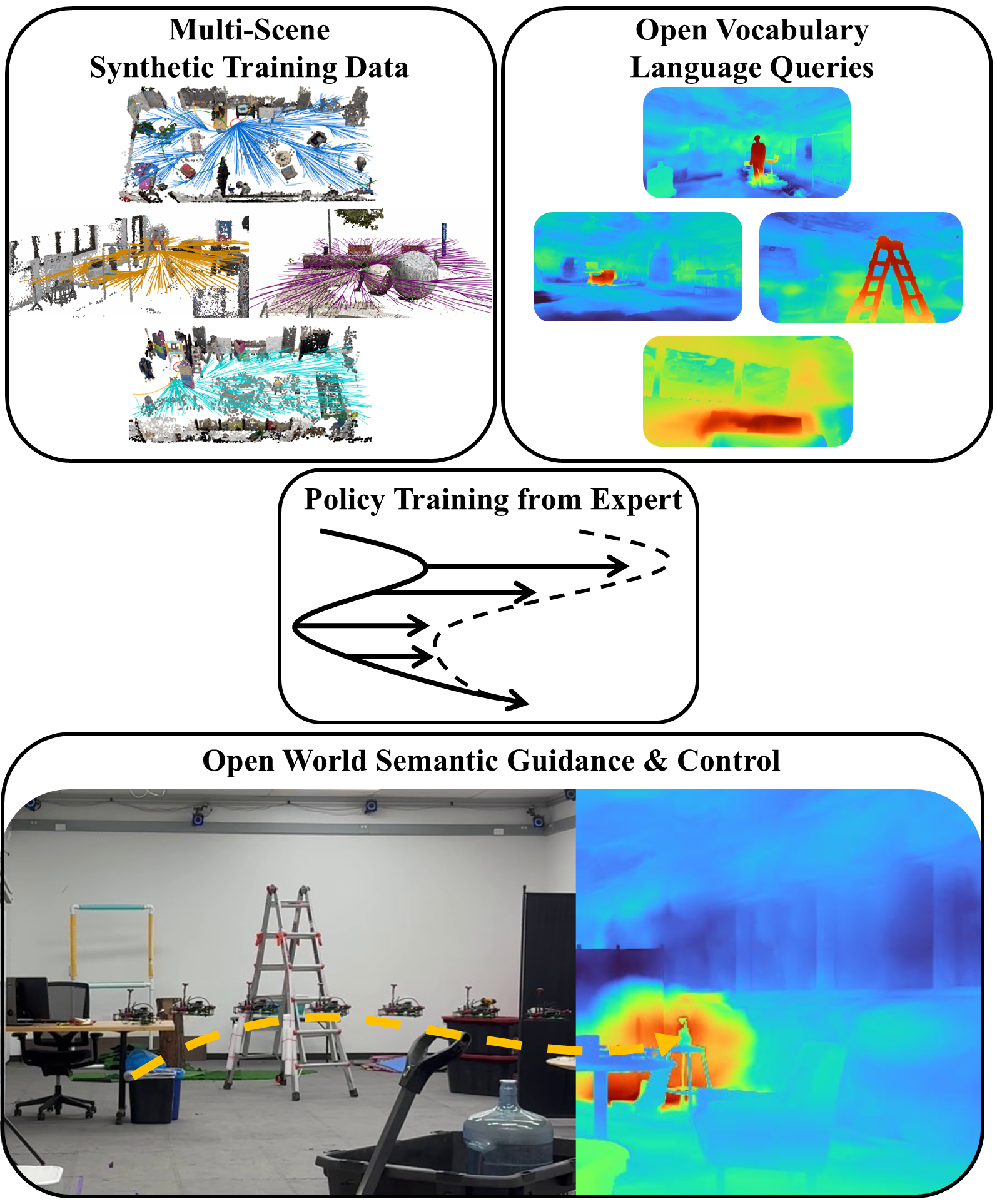}
    \caption{\textbf{SINGER}: pipeline for training and deploying open-vocabulary language conditioned guidance policies in the open world with inference entirely onboard drone hardware. Scene-spanning trajectories are generated offline using an RRT-inspired expert in a simulator with multiple different 3D Gaussian Splatting scenes. The policy is conditioned on CLIP-based semantic similarity images based on multiple different language queries rendered from an ego-view. This produces training data for imitation learning a lightweight, robust vision-language guidance and control policy.}
    \label{fig:placeholder_fig_1}
    \vspace{-5mm}
\end{figure}

Advances in diffusion policies \cite{chi2023diffusion} and vision-language-action (VLA) models \cite{zitkovich2023rt, black_0_nodate} have led to significant research breakthroughs in robot policy learning from expert demonstration via imitation, particularly in robot manipulation. 
Specifically, leveraging imitation learning on \emph{large-scale} robot manipulation datasets \cite{khazatsky2024droid, o2024open}, state-of-the-art policies endow robots with the requisite task understanding and planning capabilities necessary to perform complex tasks entirely from task descriptions provided in natural language, e.g., to ``pick up the apple and place it on a plate."
However, this paradigm has been largely unsuccessful in drone navigation, due to %
scarcity of large-scale drone navigation datasets, and effective semantic distillation methods for open-world drone navigation. This is exacerbated by inherent challenges in collecting large quantities of high quality visuomotor data on highly dynamic and naturally unstable drones.%

To address the data scarcity challenge, prior work \cite{tobin2017domain, loquercio2019deep} trains visuomotor policies for drone navigation in simulation, but the effectiveness of the resulting policies are often limited by the non-negligible sim-to-real gap. SOUS-VIDE \cite{low_sous_2025} introduces FiGS, a high-fidelity Gaussian-Splatting-based drone simulator to narrow the sim-to-real gap for stronger real-world transfer; however, FiGS lacks the semantic knowledge required for open-world drone navigation, limiting its deployment to only environments and trajectories seen during training.

In this paper, we introduce \textbf{\algname} (\textbf{S}emantic \textbf{I}n-situ \textbf{N}avigation and \textbf{G}uidance for \textbf{E}mbodied \textbf{R}obots), a pipeline for training language-conditioned drone navigation policies addressing the aforementioned limitations. \algname consists of three central components: (i) a semantics-rich photorealistic flight simulator based on 3D Gaussian Splatting for efficient data generation with expert demonstrations,
(ii) a high-level  rapidly exploring random trees (RRT*) based planner that efficiently computes spatially spanning collision-free paths to a language-specified goal by time-inverting an expanded tree, and
(iii) a robust low-level visuomotor policy that tracks the resulting high-level plans with real-time feedback.
With these components, \algname trains a lightweight viusal policy that runs onboard a drone in real-time for online navigation given a natural-language goal object.

To build an effective flight simulator, we blend the high-fidelity scene-reconstruction capabilities of Gaussian Splatting \cite{kerbl20233d} with the generalizable open-world vision-language semantic features computed by CLIP \cite{radford2021learning}, achieving minimal sim-to-real gap. 
This core design choice underpins \algname's strong zero-shot generalization capabilities to unseen tasks and environments at inference time. In particular, by abstracting goal specification to a semantic (vision-language) space, \algname effectively aligns a small dataset of synthetic expert trajectories with a broad set of tasks, yielding a data-efficient training scheme for robust visuomotor policies. We augment this training approach with domain randomization for added robustness.

At deployment, we inference CLIPSeg \cite{lueddecke22_cvpr} to produce open-vocabulary semantic images of the environment as conditioning inputs, processed by an end-to-end visuomotor drone policy for low-level drone commands.

 Through our experiments, we show that \algname outperforms baseline methods in achieving sub-meter proximity to goal by $23.33\%$ with $10\%$ less collisions and keeping the query in the field of view $16.67\%$ more often without relying on external pose estimation or map-based navigation methods.

We summarize our contributions as follows:
\begin{itemize}
    \item We introduce a high-fidelity drone simulator for efficient imitation learning in language-specified drone navigation problems built on language embedded Gaussian Splatting. 
    \item We design a RRT* trajectory planner that efficiently finds thousands of collision-free feasible trajectories across multiple Gaussian Splatting scenes and multiple semantic classes, used to produce large quantities of data for training a generalist policy.
    \item We present a real-time, lightweight, low-level visual policy architecture for language guided drone navigation using onboard sensing and compute.
    \item Using these components, we train robust visuomotor policies for drone guidance given a natural language goal specification that generalizes to never before seen environments and semantic queries.
\end{itemize}

\section{Related Work}\label{sec:related}

We review prior work in robot/drone navigation across three broad categories: (i) end-to-end \emph{vision-language action policies} that enable robots to complete tasks given user-specified natural language instructions; (ii) \emph{localization and mapping-based approaches}, which construct a map for online navigation; and (iii) \emph{modular approaches} that leverage vision-language foundation models for natural-language guidance and control.

\subsection{Vision-Language-Action Policies}\label{sec:vlapolicy} Vision-language-action (VLA) policies that leverage multimodal, cross-embodied training data have achieved remarkable performance as generalist robot policies \cite{kim2025openvla, black_0_nodate}. End-to-end VLA architectures have rapidly evolved from convolutional neural network (CNN) backbones with multilayer perceptron (MLP) action modules into more complex and sophisticated architectures such as OpenVLA \cite{kim2025openvla} that leverage pre-trained vision language models (VLM) to tokenize input queries and produce discrete action chunks. Robot foundation models such as $\pi 0$ \cite{black_0_nodate} demonstrate generality and state of the art performance in a wide variety of manipulation tasks with the addition of flow-matching. However, this success is largely restricted to the robot manipulation community, with limited demonstrations on aerial vehicle platforms due to the challenges presented by inherent instability and challenging real-time control demands of drones  \cite{wang2025uavflowcolosseorealworldbenchmark, serpiva2025racevla}. Real-time processing and accurate pose estimation remain challenging to achieve onboard self-sufficient autonomous aerial vehicles, limiting the applicability of VLA policies to UAV navigation \cite{arafat_vision-based_2023}.
Inspired by the strength of VLA policies, we design a language-conditioning technique for real-time trajectory planning and control from natural-language instructions onboard drones with constrained compute resources. 
\subsection{Localization and Mapping for Navigation}\label{sec:localmapnav}
A different paradigm for state-of-the-art language-guided vehicle navigation centers around the creation of a semantic map of the environment online, preserving the spatial information and encoding semantic information into the environment. VLMaps \cite{huang23vlmaps} includes semantic task specification for ground-robots with spatially grounded semantic objects. Finding Things in the Unknown \cite{papatheodorou_finding_2023} explores and reconstructs a target environment in real-time onboard the drone, but hardware implementation necessitates external pose-estimation to localize the drone and its map. Uncertainty aware exploration methods from semantic goals are demonstrated in SEEK \cite{Ginting2024Seek} on a quadruped and VISTA \cite{nagami2025vista} on a quadruped and drone, but these approaches are slow and reliant on offboard compute or accurate camera pose estimation, hindering performance on resource constrained systems. The applicability of these map-based approaches in aerial vehicles is limited by fragile localization and pose estimation methods such as visual-inertial-odometry (VIO) or costly LiDAR payloads to produce accurate pointclouds \cite{kondo2025dynus}. An end-to-end onboard architecture for semantic mapping and exploration has yet to be demonstrated fully onboard a drone. Our proposed method does not build a map and instead uses visual feedback from the observable environment to guide the vehicle towards a semantic query.

\subsection{Modular Natural-Language Guidance \& Control }\label{sec:natlanggnc}
Vision-language foundation models, such as CLIP \cite{radford2021learning}, have been widely used to extract visual-semantic features for downstream applications in robot manipulation \cite{pmlr-v164-shridhar22a, shen2023distilled, shorinwa2024splat}, and navigation \cite{gadre2022cow, shah2023lm, chen2025splat}.
Many existing methods, e.g.,  FAn \cite{10436161}, employ vision-language foundation models to detect entities of interest in the camera-view and use model-based control techniques to keep the query centered. Although these methods enable open-world drone navigation, these methods are generally limited to top-down tracking tasks where the object remains a fixed distance from the drone, and are too slow to run onboard a drone. Most relevant to our work, the method in \cite{quach2024gaussian} uses red or blue targets to direct the drone to turn left or right as it flies through the environment. While this approach runs fully onboard the drone, it requires a hand-tailored environment for the drone and does not employ natural language guidance, limiting its applicability in the open world. 
In contrast, our method uses imitation learning from synthetic data to train a visuomotor policy to track towards the most semantically significant object in the drone's field of view, for open-world, open-vocabulary navigation, while relying entirely on onboard sensors and compute.

Several concurrent works demonstrate similar capabilities, but all require external pose estimation or a known candidate set of semantics. Zhang et al. \cite{zhang_grounded_2025} demonstrate VLFly, a method built upon ViNT \cite{shah2023vint} for vision-language guidance and navigation, but this method relies on a-priori knowledge of what the item being searched for looks like, reducing applicability in open-world environments. GRaD-Nav++ \cite{chen2025grad} accomplishes multi-task generalization with vision-language conditioning, but is limited to three semantic queries and four flight behaviors, and can only execute combinations seen in sequences prescribed during training, in environments similar to those seen during training, reducing applicability in the open-world. Wu et al. \cite{drones9080566} also introduce a deep-reinforcement learned method for vision-guided object finding drone flight, but rely on external pose estimation in hardware experiments. Despite drawing attention to challenges posed by separation between discrete modules in image processing, learned control, and state estimation, these methods still rely on a ground-truth external pose estimate or a-priori knowledge of the environment or policy semantics to fly in the real world.
SOUS-VIDE \cite{low_sous_2025} enables training zero-shot end-to-end visuomotor drone policies through synthetic data generation for perception with 3D Gaussian Splatting (3DGS). While robust to external disturbances and small changes to the environment, this method can only learn single trajectories from human-defined waypoints in a single known environment, which is limiting in practice.

With \algname, we address all these challenges, enabling drone navigation through guidance from natural language using onboard compute and sensing.

\section{Problem Statement}
We consider a language-conditioned unmanned aerial vehicle (UAV) flight task in which a quadrotor drone is equipped with a monocular RGB camera, IMU and magnetometer, downward facing optical flow sensor for estimating altitude and velocity, and an onboard computer. 
The drone receives a user-specified open-vocabulary instruction to navigate to a desired location within an unknown environment, provide by a human operator or provided separately by an autonomous planning module.
We assume that the robot can only use onboard sensors and compute to safely navigate to the specified goal location, specifically using collective-thrust and body-rate commands only to ensure amenability to a broad swath of drone platforms.

\section{Language-Conditioned Data Synthesis}
We develop a framework to generate synthetic imitation learning data for open world UAV flight that generalizes a limited set of expert trajectories via natural language. Photorealistic synthetic camera images for policy training are generated with a semantically rich FiGS \cite{low_sous_2025} simulator. We fuse spatial and semantic embeddings in the 3DGS rendering engine to anchor simulated flight trajectories to the semantics of a set of scenes. Our simulator is composed of a lightweight drone dynamics model and a 3DGS generated using Nerfstudio \cite{nerfstudio}.

\subsection{Drone Dynamics Model}
Our model operates in the world, body, and camera frames ($\mathcal{W}$, $\mathcal{B}$, $\mathcal{C}$) and uses a 10-dimensional semi-kinematic state vector,  
$\bm{x} = \begin{bmatrix} \bm{p}_\mathcal{W},\bm{v}_\mathcal{W}, \bm{q}_\mathcal{BW} \end{bmatrix}^T$, representing position ${\bm{p}_\mathcal{W} = (p_x, p_y, p_z)}$, velocity $\bm{v}_\mathcal{W} = (v_x, v_y, v_z)$, and orientation $\bm{q}_\mathcal{BW} = (q_x, q_y, q_z, q_w)$. The control inputs,  
${\bm{u} = \begin{bmatrix} f_{th}, \bm{\omega}_\mathcal{B}\end{bmatrix}^T}$, include normalized thrust $f_{th}$ and angular velocity $\bm{\omega}_\mathcal{B} = (\omega_x, \omega_y, \omega_z)$. This constitutes the dynamics model:
\begin{equation}\label{eq:equations_of_motion}
\begin{split}
\bm{\dot{p}}_\mathcal{W} &= \bm{v}_\mathcal{W},\\[-1ex]
\bm{\dot{v}}_\mathcal{W} &= g\bm{z}_\mathcal{W} - k_{th}\frac{f_{th}}{m_{dr}} \bm{z}_\mathcal{B} \\[-1ex]
\bm{\dot{q}}_\mathcal{BW} &= \frac{1}{2} \bm{W}(\bm{\omega}_\mathcal{B})\bm{q}_\mathcal{BW},
\end{split},
\end{equation}
where $g$ is gravitational acceleration, $\bm{W}(\bm{\omega}_\mathcal{B})$ is the quaternion multiplication matrix, and $\bm{z}_\mathcal{W}$, $\bm{z}_\mathcal{B}$ are the z-axis unit vectors of the world and body frames. The thrust coefficient and mass, $(k_{th}, m_{dr})$, are stored in the drone parameter vector $\bm{\theta}$.

During synthetic expert data generation, we integrate forward the equations of motion using ACADOS \cite{Verschueren2021} to obtain the state $\mathbf{X} = \{\bm{x}_0, \dots, \bm{x}_K \}$ and input trajectory ${\mathbf{U} = \{\bm{u}_0, \dots, \bm{u}_{K-1} \}}$, where $K$ denotes the number of discrete time steps. We render the image sequence $\bm{\mathcal{I}} = \{\bm{I}_0, \dots, \bm{I}_K \}$ as seen by the onboard camera using the 3DGS.

\subsection{Semantic 3D Gaussian Splatting} The 2D vision-language model CLIP \cite{radford2021learning} is used to distill CLIP image embeddings into the 3DGS which maps a 3D point to a semantic embedding \cite{shen2023distilled, shorinwa2025siren}. This process jointly trains a scene-specific semantic field ${f: \mathbb{R}^3 \mapsto \mathbb{R}^{l}}$, parameterized by a multi-resolution hashgrid followed by a multilayer perceptron, along with the 3DGS. The semantic field may then be queried at the mean of any point in the sparse point cloud representation of the 3DGS to identify the semantics of that cluster of points. Querying for an object in the scene thus produces a point cloud representation of the object located in the frame of the 3DGS, from which we compute its 3D location. We add this functionality to FiGS \cite{low_sous_2025} to enable language-conditioned trajectory generation and collision detection.

\subsection{Spatially Spanning Trajectories:} 
\begin{figure}
    \centering
    \includegraphics[width=0.95\linewidth]{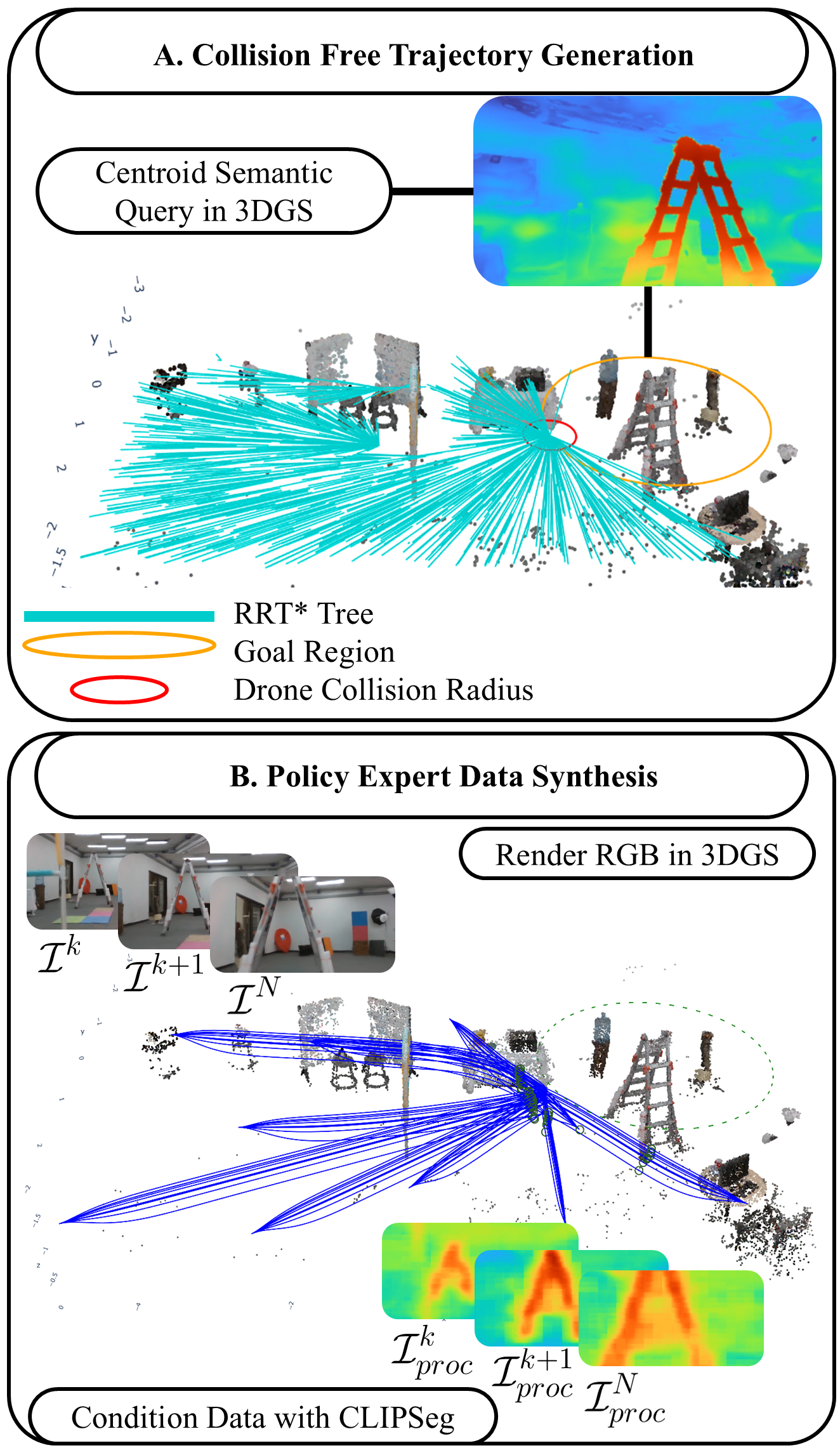}
    \caption{\textbf{The \algname data synthesis pipeline.}  (A.) Time-inverted RRT* based trajectory generation process leveraging semantic Gaussian Splatting. (B.) Natural language conditioning process applied to the policy expert data generation method.}
    \label{fig:data_synthesis}
    \vspace{-5mm}
\end{figure}

To facilitate open-world policy deployment, we use a spatially spanning trajectory generation method built on a time-inverted RRT* (Alg. \ref{alg:rrt_star_compact}). RRT* is used offline to explore the 3DGS environment spatially by random sampling free space and building branches between sampled points, or nodes $v$.
Each semantically significant object centroid $q_o\in\mathbb{R}^{3}$ in the environment is located at the root of its own RRT* tree $\mathcal{T}=\left(\mathcal{V},\mathcal{E}\right)$, and the trajectories parameterized by nodes $\mathcal{V}$ and edges $\mathcal{E}$ branch through the environment to its boundaries $\mathcal{B}$ along a horizontal (X,Y) plane located at the centroid's altitude $q_{0z}$. We create bounding bubbles around the points in the sparse point cloud and prevent the RRT* from building branches into these regions to ensure collision-free trajectories. A ``goal-region" is extended around the semantic query to influence the approach direction of the trajectory towards the center of the environment, and to prevent generating trajectories that fly the drone over furniture. RRT* performs a rewiring process to ensure each branch of the tree doesn't pass through redundant nodes. The relatively sparse RRT* branches are then segmented into trajectories in the 3D position space, and smoothed using a cubic spline. An upper bound on the drone velocity is used to generate velocity data for each branch in the RRT*. Since $q_{0}$ is known in the FiGS simulation, we use the unconstrained yaw of the drone to point the camera normal towards the semantically significant object. This process results in a coarse trajectory (Fig. \ref{fig:data_synthesis}) that is then passed to a robust optimal controller during data synthesis.
\begin{algorithm}[!b]
\caption{Sparse Point Cloud Environment-Spanning RRT*}
\label{alg:rrt_star_compact}
\begin{algorithmic}[1]
\Require Environment point-cloud $\mathcal{O}$, start $q_s$, semantic centroid $q_o$, bounds $\mathcal{B}$, step size $\eta$, iterations $N$
\Ensure Tree $\mathcal{T} = (\mathcal{V}, \mathcal{E})$

\State Initialize $\mathcal{T}$ with root $q_s$, build KD-tree $\mathcal{K}$ from $\mathcal{O}$
\State Set $\gamma \leftarrow \alpha(1 + 1/d)^{1/d}$

\For{$i = 1$ to $N$}
    \State $r \leftarrow \gamma(\log|\mathcal{V}|/|\mathcal{V}|)^{1/d}$ \Comment{Search radius}
    \State $q_{rand} \leftarrow$ uniform sample in $\mathcal{B}$
    \State $v_{near} \leftarrow$ nearest node to $q_{rand}$ in $\mathcal{T}$
    \State $q_{new}\! \leftarrow\! v_{near}\! +\! \min(\eta, \|q_{rand} - v_{near}\|)\! \cdot\! \frac{q_{rand} - v_{near}}{\|q_{rand} - v_{near}\|}$
    
    \If{$q_{new}$ violates bounds, exclusion zones, or collides}
        \State \textbf{continue}
    \EndIf
    
    \State $\mathcal{N} \leftarrow$ nodes within radius $r$ of $q_{new}$
    \State $v_{best} \leftarrow \arg\min_{v \in \mathcal{N}} \{v.cost + \|q_{new} - v\|\}$ s.t. collision-free
    \State Add $v_{new}$ to $\mathcal{T}$ with parent $v_{best}$
    
    \For{$v \in \mathcal{N} \setminus \{v_{best}\}$} \Comment{Rewire}
        \If{$v_{new}.cost + \|v - v_{new}\| < v.cost$ and collision-free}
            \State Rewire $v$ to parent $v_{new}$, update descendant costs
        \EndIf
    \EndFor
\EndFor
\end{algorithmic}
\end{algorithm}

\subsection{Policy Expert Data Synthesis}
Given the privileged information present in a simulated environment, we use a simulated robust model predictive control (MPC) expert to fly the RRT* trajectories from leaf-to-root, or inverted. We sample RGB images from the 3DGS $\mathcal{I}^{k}$, at each time step, which is processed into $\mathcal{I}_{proc}^{k}$ with CLIPSeg. The states $x_k$ and inputs $u_k$ are also stored at 20Hz from the simulated MPC flights to build our dataset. The drone's flight parameters (mass and normalized thrust coefficient) are randomized to within $30\%$ of the actual drone, and we additionally domain randomize the pose and velocity of the drone each 2s of flown trajectory. We rely on the robustness of the expert to recover from this perturbation, funneling the drone towards the nominal trajectory similar to tube-based MPC \cite{tagliabue2024tube}. This builds a robust dataset of state and input data enveloping a wide distribution of possible sources of modeling error, including drone configurations and flight conditions, and has been demonstrated to imbue the learned policy with robustness to low battery, poorly estimated mass, rotor downwash, and rough wind conditions \cite{low_sous_2025}. All sensor measurements and output states and actions for each trajectory are split into these 2s segments and shuffled, comprising the training data samples used to train the policy end-to-end.
We use semantically segmented image data generated with CLIPSeg \cite{lueddecke22_cvpr} instead of unprocessed RGB images. The output logits are mapped to a perceptually uniform 3-channel colormap, effectively transforming the images that the policy sees to a semantic-spatially aligned space (Fig. \ref{fig:data_synthesis}). In doing so, we generalize the policy across environments on the basis of semantics. This generalized environment representation is a ``semantic heatmap" where regions of the environment more semantically similar to the user provided query are brighter red, and less semantically similar regions are dark blue. When left unnormalized, the semantic query is bright red or orange, most of the environment is yellow or green, and the least semantically relevant parts of the environment are blue.

\section{\algname Policy Architecture and Training}
\begin{figure}
    \vspace{2mm}
    \centering
    \includegraphics[width=1.0\linewidth]{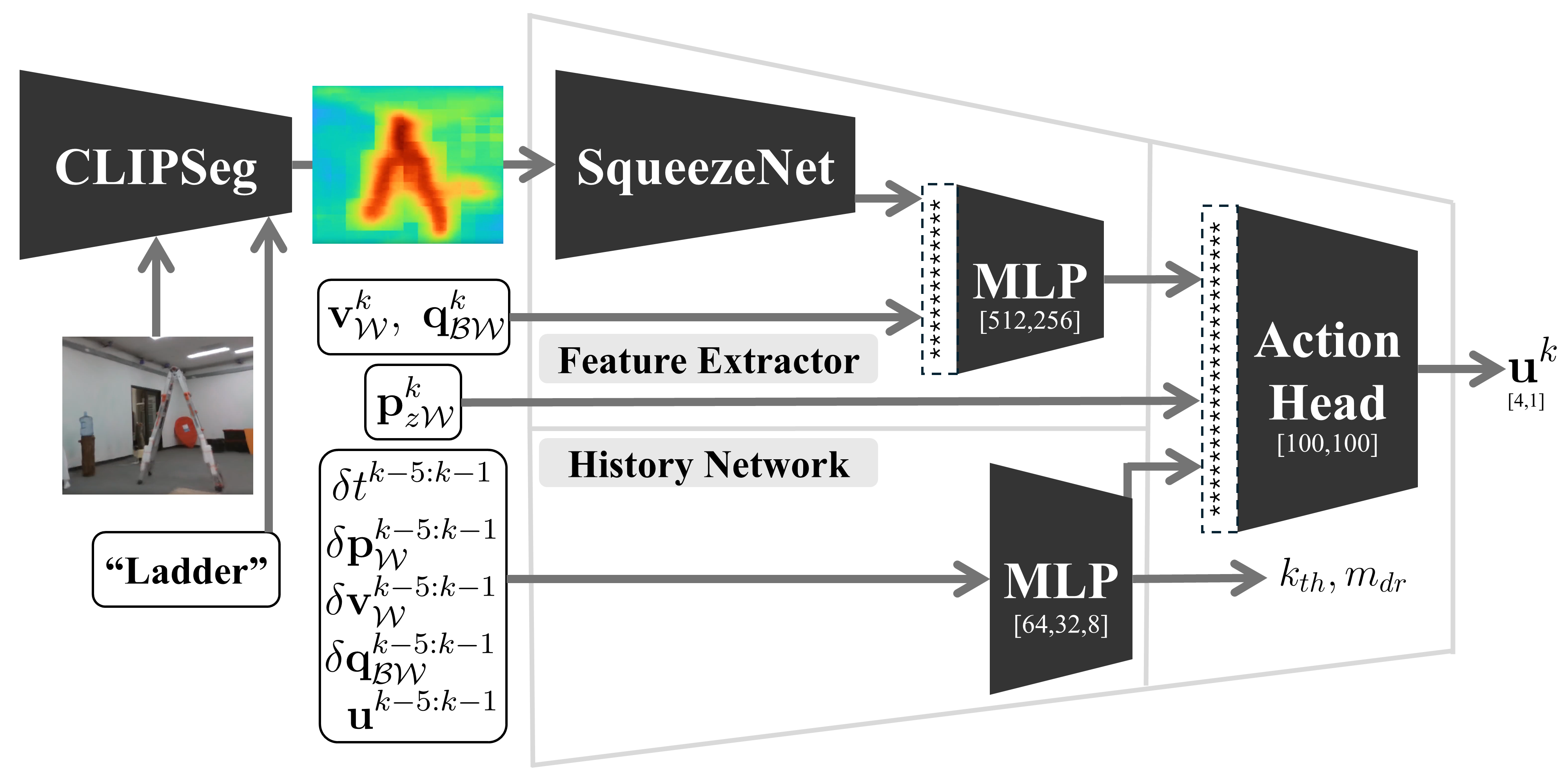}
    \caption{\textbf{Training architecture.} \algname is trained on images consisting of the output logits from CLIPSeg, a pixel-dense semantic segmentation network built on CLIP. At runtime, a semantic query is passed into CLIPSeg along with monocular RGB images. The output patch-logit images are then ingested by the feature extractor network   to produce body-rate motor commands. Network dimensions are listed for the feature extractor, history network, and action head.}
    \label{fig:ssv-net}
    \vspace{-5mm}
\end{figure}

The deep learned policy architecture is adopted from the SV-Net described in \cite{low_sous_2025}, with an 
additional image pre-processing step appended to the feature extractor network. Instead of training the policy 
on raw RGB camera images, we process these images with CLIPSeg \cite{lueddecke22_cvpr} and train on the 
image-space of CLIPSeg logits. When trained in this way, we refer to the trained policy as 
\algname (Fig. \ref{fig:ssv-net}). The two-stage training procedure prescribed in \cite{low_sous_2025} is used to 
first train a history network to predict time-varying system parameters in a latent vector by ingesting a sliding 
window of changes in observable states. This network is trained with a loss on the true mass and thrust coefficient 
of the drone, which can capture changes in the dynamics model within the distribution of domain randomization 
applied during data generation. The feature extractor and action head are trained end-to-end once the 
history network weights are frozen. This full network is trained with a loss on the expert demonstrator's 
motor commands over the 2s trajectory chunks. We pass the output patch-logits from CLIPSeg into the feature 
extractor along with the current state measurement during training.

\algname produces motor commands at 20Hz, but CLIPSeg uses CLIP based on the ViT-B/16 vision transformer model 
with 86M parameters. This imposes a significant bottleneck on the inference time of the policy (3Hz on NVIDIA 
Jetson Orin Nano 8Gb), and is the primary reason why similar works stream motor commands to the drone from a 
separate workstation or laptop. In contrast, we instead trace CLIPSeg with the Open Neural Network Exchange (ONNX) 
neural network interoperability standard to produce a lightweight computational graph, and inference using the 
CUDA ONNX runtime. This facilitates inference at up to 12Hz onboard the NVIDIA Jetson Orin Nano 8GB, which we do 
asynchronously from 20Hz \algname inference.

At policy inference, we normalize the semantic similarity score against the highest score seen during flight. This gives 
the policy a rudimentary memory of what it's seen, so that it can only guide itself towards more semantically similar regions in 
its field of view.

We generate approximately 1650 trajectories across 15 different semantic queries in 5 different 3DGS environments representing both indoor and outdoor environments. Amounting to 907,440 samples of observation-action labeled data pairs, this data captures a finite range of environments and objects that the drone might see in the wild. We train the SV-Net on this dataset using a workstation with an NVIDIA RTX 4090 and an Intel i9-14700.

\section{Experiments}
We evaluate the performance of the \algname in drone experiments to evaluate its generalization and robustness capabilities in simulation within a 3DGS environment and in the real world on hardware. These experiments comprise the policy being provided with a semantic query, and needing to fly up to a goal-region extending 2.0m from the query as done in policy training.

\subsection{Simulated Experiments - \algname in Simulation}
The performance of \algname is evaluated in three simulated scenarios designed to test the generalization of the
approach to new semantic queries and environments. The easiest scenario corresponds to a 3DGS environmnent and 
semantic queries from the training distribution. The intermediate scenario corresponds to a selection of three 
3DGS environments used during policy training, with semantic queries seen during training, but not in the environments 
tested here. Finally, the hardest scenario is designed to evaluate policy performance in a unseen environment 
and on unseen semantic queries.

\subsection{Baseline and \algname On Hardware}
We evaluate the real-world performance of \algname against a baseline in six hardware experiments with five trials each, corresponding to three semantic queries with two initial locations each in a close-quarters mockup office environment. None of the objects corresponding to the semantic queries, nor the environment itself were seen during policy training. %
The semantic query is in-view at the beginning of the experiment. The policy is evaluated on successful flight towards the queried object without collisions.
We compare \algname to a classical guidance and control implementation relying on CLIPSeg to process RGB images from the onboard camera. The baseline is most similar to \cite{10436161} in implementation and \cite{quach2024gaussian} in deployment. The baseline centers the image masked by CLIPSeg in the field of view of the camera onboard the drone while flying at a fixed velocity towards it until the mask occupies the majority of the image. A simple PD controller tracks an orientation set point designed to keep the mask centered in the camera view, and a constant along-track velocity is applied. This baseline constitutes a minimal implementation of open-vocabulary vision-language UAV guidance. In these experiments, true-north of the world frame is provided externally to prevent uncertain and varying compass heading measurements from affecting the experimental outcome. We additionally compare against \algname with no reliance on motion capture in the same environment to demonstrate fully onboard implementation, although this experiment is subject to magnetometer measurement error.

\subsection{Hardware Platform}\label{sec:hardware}
We deploy this system on a 5 inch Lumenier Cine-whoop drone equipped with a Pixracer R15 Pro, NVIDIA Jetson Orin Nano, ZED Mini camera, and ARK Flow optical flow and rangefinder. The Zed Mini camera is used as a monocular RGB camera in all experiments. All policy inference is onboard the drone in the \algname policy implementations and baseline. Due to computational bottlenecks, the ZED Mini is operating at the lowest available resolution (VGA $672\times376$px) to reduce image processing overhead. The CLIPSeg ViT-B/16 patch size is 16$\times$16, corresponding to 42$\times$24 patches of input for the camera image stream. The image processing pipeline runs at $12-13$Hz, the majority of which is the forward pass through CLIPSeg. 

\subsection{Results}\label{sec:results}
\begin{figure}[h]
\vspace{1mm}
    \centering
    \includegraphics[width=1.0\linewidth]{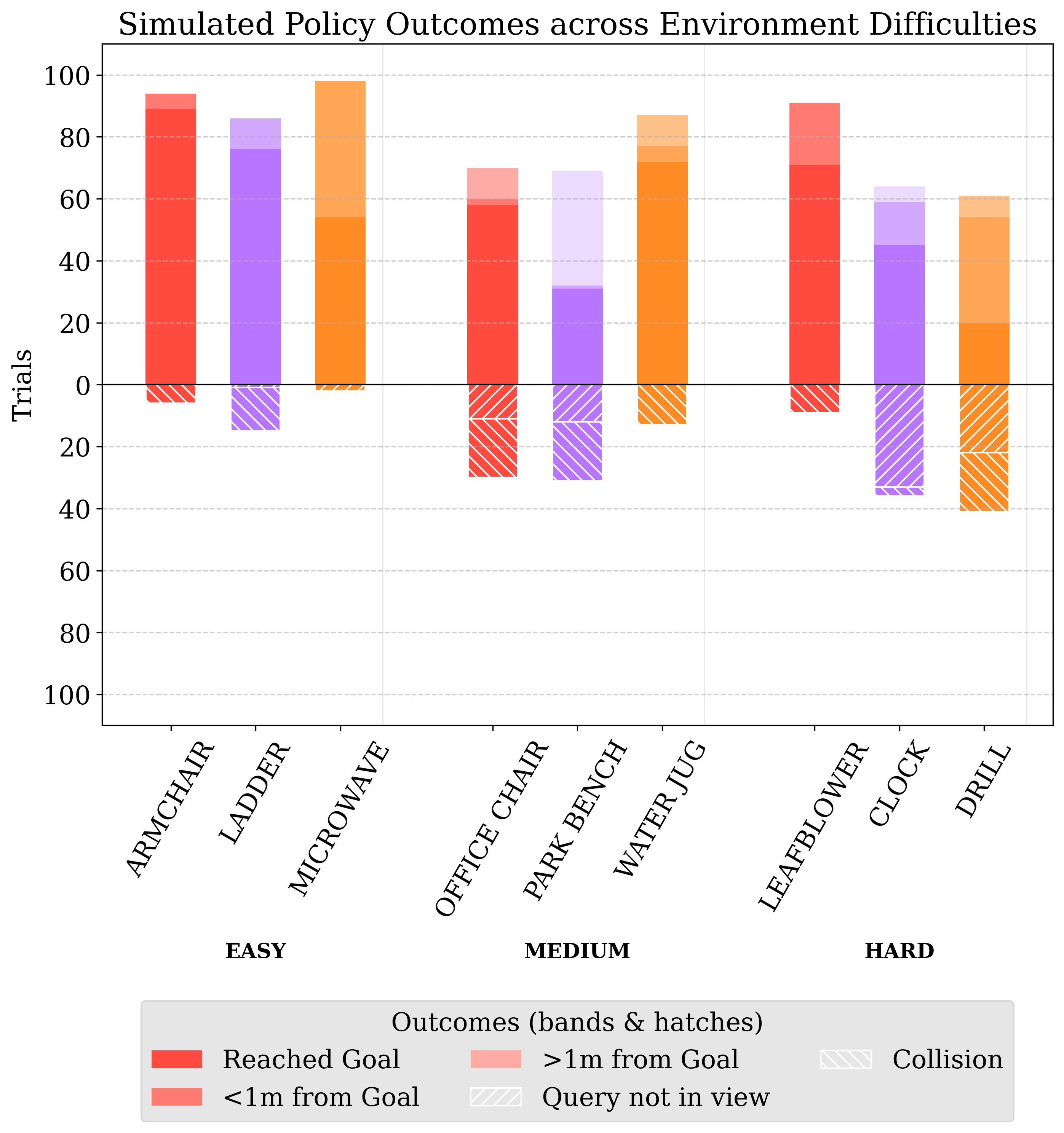}
    \caption{\textbf{\algname evaluated in simulation across three different 3DGS environments.} The darkest of each colored bar denotes reaching the goal, while the middle hue shows $<1m$ from the goal, and the lightest color indicates policy stopping $>1m$ from the goal. Failures are shown below zero, with crosshatch directions denoting cause of failure. Collisions are counted while the policy has the query in-view, while query-not-in-view describes cases where the drone did not fly towards the prescribed query. EASY is a fully-in-distribution test in an environment used to train the policy. MEDIUM is an in-distribution environment with in-distribution queries not trained on in the tested environment. HARD is a fully out of distribution environment and semantic queries.}
    \label{fig:singer_exp_sim}
    \vspace{-5mm}
\end{figure}

\algname was tested in simulation in 90 experiments with 10 trials each across six different environments and nine semantic queries.
Each experiment consisted of a randomized initial location in the 3DGS environment and a semantic query, and ten policy rollouts.
As shown in Fig.~\ref{fig:singer_exp_sim}, the easy environment and semantic queries were drawn directly from the training distribution. \algname performs the 
best at this experiment difficulty, reaching the goal region $73\%$ of the time, and reaching sub-meter proximity $92.7\%$ of the time with minor failures in some trials due to collisions 
or never seeing the correct semantic query. The medium set of experiments introduced three unique combinations of scenes and semantic 
queries. Each scene was drawn from the training distribution, as were the semantic queries, but the combinations of scene and query tested here were 
not seen during training. The policy performs particularly poorly on the park bench query. The environment this query was tested in
has many initial starting locations for the drone where the semantic query is completely occluded. In these cases, it is less likely for the 
policy to identify the semantic query, and it may fly towards the most semantically similar object in its field of view, in a path that 
may not take it towards the true semantic query. Finally, \algname was tested in a fully unseen environment with semantic queries unseen during 
training. This environment has the lowest overall rate of reaching the goal region, and the most collisions, including collisions due to not seeing the correct query.

When deployed in hardware in the hardest evaluation scenario (three unseen semantic queries in an unseen deployment 
environment) \algname performs the best overall, keeping all semantic queries in view during flight and getting within $<1$m 
of the goal region $76.67\%$ of the time ($23/30$ trials), and stopping outside of this range $6.67\%$ of the time ($2/30$). Our policy only collides
with obstacles in the environment $16.67\%$ of the time ($5/30$). The baseline performs comparably or worse across all semantic queries 
getting within $<1$m of the goal $53.33\%$ of the time ($16/30$), stopping short $3.33\%$ of the time ($1/30$), and colliding with the environment 
$26.67\%$ of the time ($8/30$). The baseline fails to track the correct semantic query $16.67\%$ of the time ($5/30$), demonstrating the limited semantic scene understanding of the baseline compared to \algname.
Crucially, \algname always finds the right semantic query in its field of view, and only fails when its path takes it too close to 
the obstacles in the scene. Both the baseline and \algname use the same output patch logits from CLIPSeg, but the baseline relies on segmentation and 
centroiding, which can be unreliable across multiple frames as the point of view of the drone changes as it moves through the environment.
When the external true-north is removed from \algname and it must rely on its internal sensors, \algname still performs 
comparably or better than the baseline, reaching sub-meter distance from the goal $66.67\%$ ($20/30$), and colliding with the environment the same proportion ($16.67\%$) of the time ($5/30$).
Without a reliable true-north, the onboard magnetometer is susceptible to varying external magnetic fields induced by heavy machinery nearby.
This results in one more failure case ($6/30$)  vs. the baseline at $(5/30)$ due to tracking the incorrect semantic query, as the drone cannot maintain a good estimate of its orientation.
The baseline was completely unable to perform without an externally provided true north heading, as the velocity set point requires a reliable heading in the world frame. Without this, the baseline would fly in arbitrary directions as the magnetic field switched direction during flight, changing the world-frame reference being used by the flight controller. We include \algname's results under the same conditions as a testament to its ability to outperform the baseline.
Moreover, the overall policy performance is comparable to that in simulation, highlighting the efficacy by which our simulator reduces the sim-to-real gap in perception.

\begin{figure}
\vspace{1mm}
    \centering
    \includegraphics[width=1.00\linewidth]{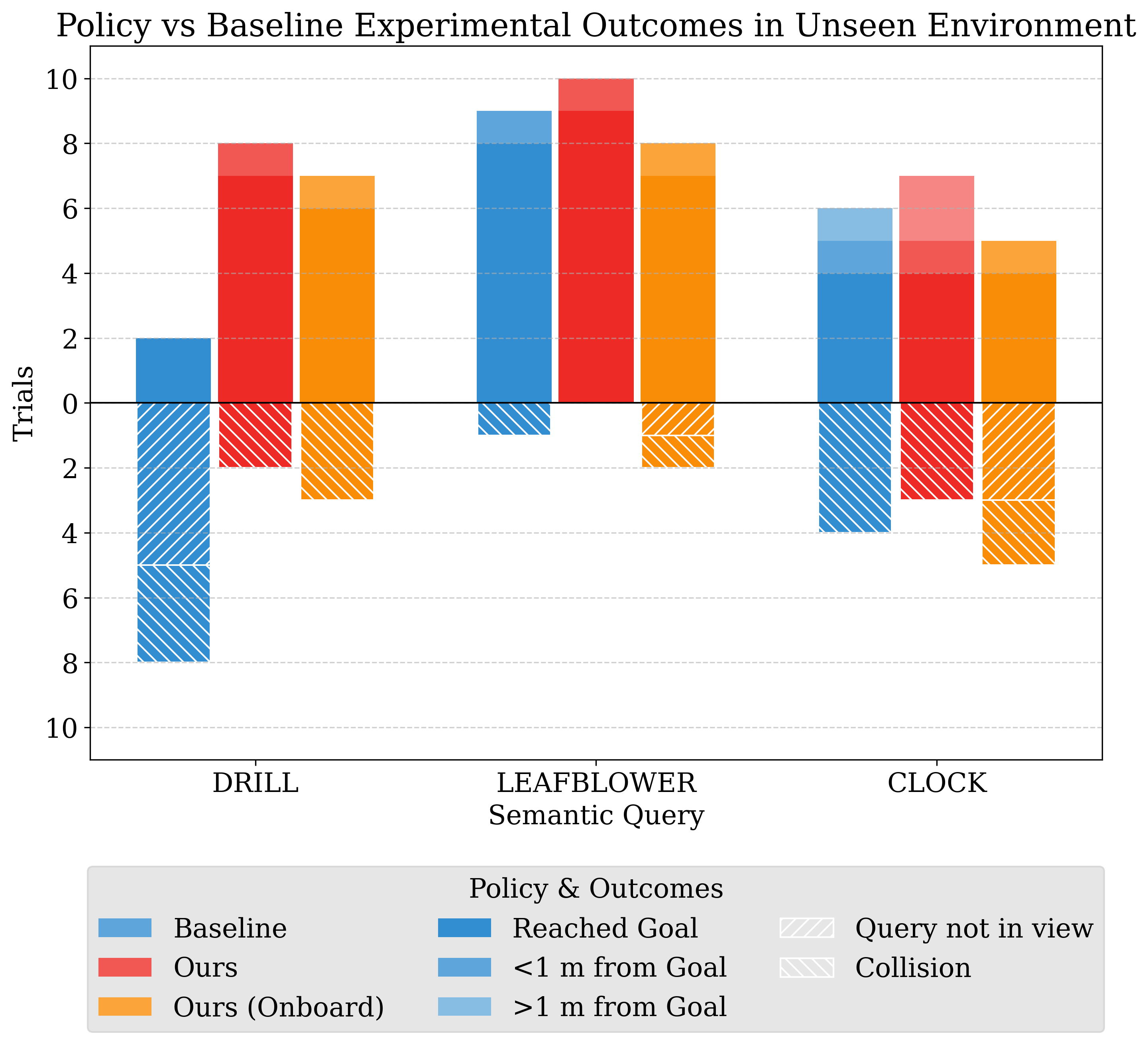}
    \caption{\textbf{Experimental results comparing \algname to a yaw-rate PD controlled baseline, and to \algname with fully 
    onboard sensors.} Our policy performs the best given access to the same information as the baseline, never failing to maintain 
    sight of the semantic query, and consistently reaching the goal region. All trials are depicted, with successful trials 
    above unsuccessful ones. Crosshatching direction on unsuccessful trials denotes the reason for failure, where collisions are counted while the policy has the query in-view, while query-not-in-view describes cases where the drone did not fly towards the prescribed query.}
    \label{fig:experimental_results}
    \vspace{-5mm}
\end{figure}

The most challenging semantic query tested in hardware was the clock, which \algname performed the worst on in simulation. The overall success rate of the policy in-simulation is also comparable to the results in hardware.
The main limitations of all tested policies largely come from lack of collision avoidance and reliance on low-resolution camera images. We found that the policy performs poorly when the semantic query is poorly resolved, as CLIPSeg is not able to reliably segment it frame-to-frame. 

\section{Conclusion}
In this work, we have presented a novel technique for achieving generalizable language-conditioned autonomous drone navigation with fully-onboard sensing and compute using a lightweight visuomotor policy trained using imitation learning on natural-language guided trajectories. We improve upon the single-trajectory, single-environment limitation of existing policies
through careful curation of language-embedded data and image pre-processing. However, we also recognize collision avoidance as a new challenge introduced by deployment in the open-world and highlight this as an impactful open area for future work. Additionally, the low-level (guidance and control) implementation of our method limits capabilities to ``query-seeking" behavior. Conversely, this policy implementation highlights the modularity of \algname. Natural language guidance and control opens the door for integration with hierarchical navigation methods that provide sequential semantic queries. Through rigorous testing, we demonstrate that \algname can be used to guide drones towards observed semantic queries using onboard sensing and compute. This method is amenable to drones with a front-facing monocular camera, IMU, altimeter, and velocity estimation. Directions for future work include policy conditioning on more verbose queries, maneuvers, environmental interaction, and responsiveness to dynamic environments.
\vspace{5mm}

\vspace{-5mm}
\section*{ACKNOWLEDGMENT}
We acknowledge the use of generative AI tools exclusively for tab-auto-completion while extending handwritten code for plotting to new scripts while and debugging (Copilot/ChatGPT).
\vspace{-5mm}

\bibliographystyle{IEEEtran}
\bibliography{bibliography}

\end{document}